\documentclass[11pt,a4paper]{article}
\usepackage[usenames,dvipsnames,svgnames,table]{xcolor}
\usepackage[hyperref]{naaclhlt2018}
\usepackage{times}
\usepackage{latexsym}

\usepackage{sutton_math_symbols}
\usepackage{amssymb}
\usepackage{amsmath}
\usepackage{amsfonts}
\usepackage{breqn}
\usepackage{multirow}

\newcommand{\alphaB}{\boldsymbol{\alpha}}
\newcommand{\betaB}{\boldsymbol{\beta}}

\newcommand{\cscomment}[1]{}
\newcommand{\comment}[1]{}
\newcommand{\cs}[1]{\comment{\textcolor{Purple}{\textit{\lbrack CAS: #1 \rbrack}}}}
\newcommand{\as}[1]{\comment{\textcolor{BrickRed}{\textit{\lbrack AS: #1 \rbrack}}}}
\usepackage{xspace}

\newcommand{\dPAM}{{aviPAM}\xspace}

\usepackage{url}
\usepackage[]{algorithm2e}
\usepackage{graphicx} 
\usepackage{subfigure} 

\aclfinalcopy 

\title{Variational Inference In Pachinko Allocation Machines}

\author{Akash Srivastava \\
  University of Edinburgh  \\
  {\tt akash.srivastava@ed.ac.uk} \\\And
  Charles Sutton \\
  University of Edinburgh \\
  The Alan Turing Institute \\
  Google Brain \\
  {\tt csutton@inf.ed.ac.uk} \\}

\date{}

\begin{document}
\maketitle
\begin{abstract}
The Pachinko Allocation Machine (PAM) is a deep topic model that allows
representing rich correlation structures among topics by a 
directed acyclic graph over topics. Because of the flexibility
of the model, however, approximate inference is very difficult.
Perhaps for this reason, only a small number of potential PAM architectures
have been explored in the literature.
In this paper we present an efficient and flexible amortized variational 
inference method for PAM, using a deep inference network to parameterize
the approximate posterior distribution in a manner similar to the variational autoencoder.
Our inference method produces more coherent topics than state-of-art inference methods for PAM
while being an order of magnitude faster,
 which allows exploration of a wider range of PAM architectures than have previously been studied.
\end{abstract}

\section{Introduction}
\label{intro}

Topic models are widely used tools for exploring and visualizing document collections. Simpler topic models, like latent Dirichlet allocation (LDA) \citep{blei2003latent}, capture correlations among words
but do not capture correlations among topics.
This limits the model's ability to discover finer-grained
hierarchical latent structure in the data. For example, we expect that 
very specific topics, such as those pertaining to 
individual sports teams, are likely to co-occur more often
than more general topics, such as a generic ``politics'' topic
with a generic ``sports'' topic.

A popular extension to LDA that captures topic correlations is 
the Pachinko Allocation Machine (PAM) \citep{li2006pachinko}.
PAM is essentially ``deep LDA''.
It is defined by a directed acyclic graph (DAG) in which each leaf node denotes a word in the
vocabulary, and each internal node is associated with a distribution
over its children. The document is generated by sampling, for each word, a path from the root of the DAG to a leaf.
Thus the internal nodes can represent distributions over topics, so-called ``super-topics'' that represent correlations among topics.

Unfortunately PAM introduces many latent variables --- for each word in the document, the path in the DAG that generated the word
is latent. Therefore, traditional inference methods, such as Gibbs sampling and decoupled mean-field variational inference, become significantly more
expensive.
This not only affects the scale of data sets that can be considered,
but more fundamentally the computational cost of inference makes it difficult to explore the space of possible architectures for PAM. As a result,
to date only relatively simple architectures have been studied in the literature \cite{li2006pachinko,mimno2007mixtures,li2012nonparametric}.

We present what is, to the best of our knowledge, the first variational inference method for PAM,
which we call \textit{Amortized Variational Inference for PAM} (\dPAM). 
Unlike collapsed Gibbs, \dPAM can be generically applied to any PAM architecture  without the need to derive a
new inference algorithm,
allowing much more rapid exploration of the space of possible model architectures.

\dPAM is an \emph{amortized inference} method that follows the learning principle of variational autoencoders (VAE) \citep{kingma2013auto,rezende2014stochastic},
which means that all the variational distributions are parameterized by deep neural networks (encoder/inference-network) that are trained
to perform inference. The actual observation model in such a framework is often referred to as the decoder. \dPAM introduces a novel structured VAE since the existing VAE architectures cannot deal with the highly complicated latent spaces of PAMs.
We find that \dPAM is not only an order of magnitude faster than collapsed Gibbs,
but even returns topics with greater or comparable coherence. 
The dramatic speedup in inference time comes from the \textit{complete} amortization of the learning cost via our highly structured encoder architecture (neural network) that directly outputs all the variational parameters of the approximate posterior over all the latent variables in PAM,
instead of learning them separately for each training instance. 
This efficiency in inference enables exploration of more complex and deeper PAM models than have previously been possible. 
 
As a demonstration of this,
as our second contribution we introduce a mixture of PAMs model. By mixing PAMs with varying numbers of topics, this model captures the latent structure in the data at many different levels of granularity that decouples general broad topics from the more specific ones.

Like other variational autoencoders, our model also suffers from the problem of posterior collapse \citep{van2017neural},
which is sometimes also called component collapse \citep{dinh2016training}. 
We present an analysis of these issues in the context of topic modeling and propose a normalization based solution to alleviate them.

\section{Latent Dirichlet Allocation}
\label{dlda}

LDA represents each document $\wB$ in a collection as an admixture of topics.
Each topic vector $\beta_k$ is a distribution over the vocabulary, that is, a vector of probabilities, and $\beta = (\beta_1 \ldots \beta_K)$ is the matrix of the $K$ topics.
Every document is then generated under the model by first sampling
 a proportion vector $\theta \sim \mbox{Dirichlet}(\alpha)$, and then for each word at position $n$, sampling a topic indicator $z_n \in \{1,\ldots K\}$
as $z_n \sim \mbox{Categorical}(\theta)$, and finally sampling the word index $w_n \sim \mbox{Categorical}(\beta_{z_n})$.

\cscomment{
The marginal likelihood of a document $\wB$ is therefore
\begin{multline}
\label{eq:lda_1}
p(\wB|\alpha,\beta) = \\ \int_\theta \left( \prod_{n=1}^{|\wB|} \sum_{z_n = 1}^k p(w_n|z_n,\beta) p(z_n|\theta)\right) p(\theta|\alpha)d \theta.
\end{multline}
}


\subsection{Deep LDA: Pachinko Allocation Machine}
\label{dpam}

PAM is a class of topic models that extends LDA by modeling correlations among topics. 
A particular instance of a PAM represents the correlation structure among topics by a DAG
in which the leaf nodes represent words in the vocabulary and the internal nodes represent topics.
Each node $s$ in the DAG is associated with a distribution $\theta_s$ over its children,
which has a Dirichlet prior. There is no need to differentiate between nodes in the graph and the distributions $\theta_s$,
so we will simply take $\{\theta_s\} 
\cup \{1\ldots V\}$ to be the node set of the graph,
where $V$ is the size of the vocabulary.  To generate a document in PAM, for each word we sample a path 
from the root to a leaf, and output the word associated with that leaf.

More formally, we present the special case of 4-PAM,
in which the DAG is a $4$-partite graph.\footnote{An $\ell$-partite graph is the
natural generalization of a bipartite graph.} It will be clear how to generalize this discussion to arbitrary DAGs.
In 4-PAM, the DAG consists of a root node $\theta_r$ which is connected to children $\theta_1 \ldots \theta_S$ 
called \emph{super-topics}. Each super-topic $\theta_s$ is connected to the same set of children $\beta_1 \ldots \beta_K$
called subtopics, each of which are fully connected to the vocabulary items $1 \ldots V$ in the leaves.

A document is generated in 4-PAM as follows. First, a single matrix of subtopics $\beta$ are generated for the
entire corpus as $\beta_k \sim \mbox{Dirichlet}(\alpha_0)$.
Then, to sample a document $\wB$, we sample child distributions for each remaining internal node in the DAG.
For the root node, $\theta_{r}$ is drawn from a Dirichlet prior $\theta_{r} \sim \mbox{Dirichlet}(\alpha_r)$, and similarly for each super-topic $s \in \{1\ldots S\}$, the supertopic $\theta_{s}$ is drawn as $\theta_s \sim \mbox{Dirichlet}(\alpha_s)$.
Finally, for each word $w_n$, a path is sampled from the root to the leaf. From the root, we sample the index of a supertopic $z_{n0} \in \{1 \ldots S\}$ as $z_{n0} \sim \mbox{Categorical}(\theta_r)$, followed by a subtopic index $z_{n2} \in \{ 1 \ldots K\}$ sampled as $z_{n2} \sim \mbox{Categorical}(\beta_{z_{n1}})$, and finally the word is sampled as
$w_n \sim \mbox{Categorical}(\beta_{z_{n1}})$.
This process can be written as a density
\begin{align}
	P(\wB, \zB, \theta &\,|\, \alphaB, \betaB) = p(\theta_r | \alpha_r) 
 \prod_{s=1}^S P(\theta_{s} |\alpha_s)
\label{eq:dpam}\\ &\times \prod_{n} p(z_{n1}|\theta_r) 
p(z_{n2} |\theta_{z_{n1}})p(w_n|\beta_{z_{n2}}). \nonumber
\end{align}
It should be easily seen how this process can be extended to arbitrary $\ell$-partite graphs,
yielding the $\ell$-PAM model, and also to arbitrary DAGs.
Observe also that in this nomenclature, LDA exactly corresponds to 3-PAM.


\section{Mixture of PAMs}
\begin{figure}[h]
\centering
        \includegraphics[width=0.4\textwidth]{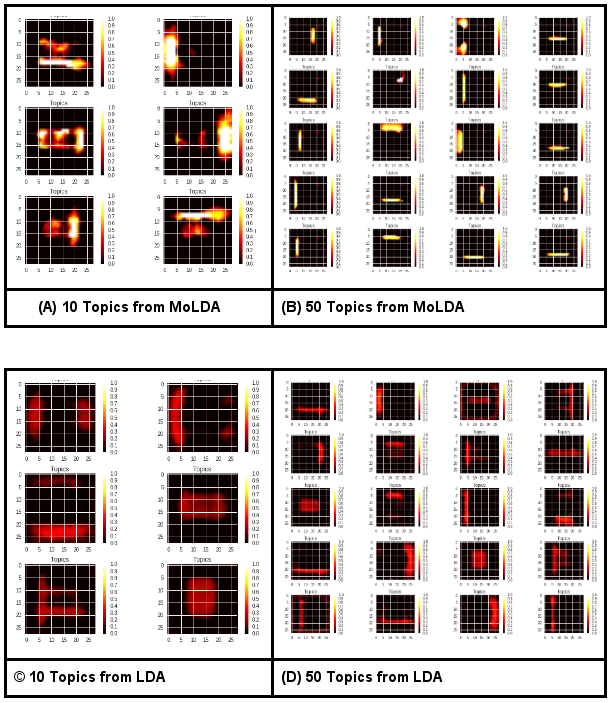}
    \caption{Top: A and B show randomly sampled topics from MoLDA(10:50). Bottom: C and D
    show randomly sampled topics from LDA with 10 topics and 50 topics on Omniglot. Notice that by using a mixture, the MoLDA can decouple the higher level structure (A) from the lower-level details(B).}
    \label{fig:mix}
\end{figure}

The main advantage of the inference framework we propose is that it allows easily
exploring the design space of possible structures for PAM.
As a demonstration of this, we present a word-level mixture of PAMs that allows learning finer grained topics
than a single PAM, as some mixture components learn topics that capture the more general, global topics
so that other mixture components can focus on finer-grained topics.

We describe a word-level mixture of $M$ PAMs $P_1 \ldots P_M$, each of which can have a different number of topics or even a 
completely different DAG structure.
To generate a document 
under this model, 
first we sample an $M$-dimensional document level mixing proportion $\theta_{r} \sim \mbox{Dirichlet}(\alpha_r)$.
Then, for each word $w_n$ in the document, we choose one of the PAM models by sampling
 $m \sim \mbox{Categorical}(\theta_r)$ and then finally sample a word by sampling a path through $P_m$ as described in the previous section. 
This model can be expressed as a general PAM model in which the root node $\theta_r$ is connected to the root nodes
of each of the $M$ mixture components.
If each of the mixture components are 3-PAM models, that is LDA, then we call
the resulting model a mixture of LDA models (MoLDA).\footnote{It would perhaps be more proper to call this model an \emph{admixture} of LDA models.}

The advantage of this model is that if we choose to incorporate different mixture components with different numbers
of topics, we find that the components with fewer topics explain the coarse-grained structure in the data,
freeing up the other components to learn finer grained topics.
For example,
the Omniglot dataset contains 28x28 images of handwritten alphabets from artificial scripts. 
In Figure \ref{fig:mix},
panels (C) and (D) are visualization of the latent topics that are generated using vanilla LDA with 10 and 50 topics, respectively. 
Because we are modelling image data, each topic can also be visualized as an image.
Panels (A) and (B) show the topics from a single MoLDA with two components, one with 10 topics and one with 50 topics.
It is apparent that the MoLDA topics are sharper, indicating that each individual topic is capturing
more detailed information about the data. 
The mixture model allows the two LDAs being mixed to focus exclusively on higher (for 10 topics) and lower (for 50 topics) level features while modeling the images. Since the final image is modeled by mixing these topics, such a mixture model with extremely sharp topics will lead to a sharper image with detailed features.
On the other hand, the topics in the vanilla LDA need to account for all the variability in the dataset using just 10 (or 50) topics and therefore are fuzzier. This in turn leads to blurry images when the topics (from (c) or (d)) are mixed to generate the images.    

\section{Inference}
\label{inference}

Probabilistic inference in topic models is the task of computing posterior distributions 
$p(\zB | \wB, \alphaB, \betaB)$ over the topic assignments for words,
or over the posterior $p(\theta | \wB, \alphaB, \betaB)$ of topic proportions for documents.
For all practical topic models, this task is
 intractable. Commonly used methods include Gibbs sampling \citep{li2006pachinko,blei2004htm}, which can be slow to converge, and variational inference methods 
such 
as mean field \citep{blei2003latent,blei2007correlated}, 
which sometimes sacrifice topic quality for computational efficiency. 
More fundamentally, 
these families of approximate inference algorithms tend to be 
model specific and require extensive mathematical sophistication on the practitioner's part since even the slightest changes in model assumptions may require substantial adjustments to the inference. 
The time required to derive new approximate inference algorithms dramatically
slows explorations through the space of possible models.

In this work we describe a generic, amortized approximate 
inference method \dPAM for learning in the PAM family of models, that is extremely fast, flexible and accurate. 
The inference method is flexible in the sense that it can be generically applied to any DAG  structure for PAM,
without the need to derive a new variational update.
The main idea is that we will approximate the posterior distribution $p(\theta_s | \wB, \alphaB, \betaB)$ for
each super-topic $\theta_s$ by a \emph{variational distribution} $q(\theta_s | \wB)$.
Unlike standard mean field approaches, in which $q(\theta_s | \wB)$ has an independent set of variational parameters
for each document in the corpus, the parameters of $q(\theta_s | \wB)$ will be computed by an 
\emph{inference network}, which is a neural network that takes the document $\wB$ as input,
and outputs the parameters of the variational distribution. 
This is motivated by the observation that similar documents can be described well by similar posterior parameters.

In \dPAM, we seek to approximate the posterior distribution $P(\theta |  \wB, \alphaB, \betaB)$, that
is, the paths $z_n$ for each word are integrated out.
Note that this is in contrast to previous collapsed Gibbs methods for PAM \cite{li2006pachinko},
which integrate out $\theta$ using conjugacy.
To simplify notation, we will describe \dPAM for the special case of 4-PAM,
but it will be clear how to generalize this discussion to arbitrary DAGs.
So for 4-PAM, we have $\theta = (\theta_r, \theta_1 \ldots \theta_S)$.

We introduce a variational distribution $q(\theta |  \wB) = q(\theta_r | \wB) q(\theta_1| \wB) \ldots q(\theta_S | \wB)$.

To choose the best approximation $q(\theta|\wB),$ we construct a lower bound to the evidence (ELBO)
using Jensen's inequality, as is standard in variational inference. 
For example, the log-likelihood function $\log p(\wB| \alpha, \beta)$ 
for the 4-PAM model \eqref{eq:dpam} can be lower bounded by  
\begin{equation}
\begin{split}
\label{eq:elbo}
\mathcal{L} =  &-\text{KL}[q(\theta_r|\wB)||p(\theta_r|\alpha_r)]\\
&- \sum_{s=1}^S
\text{KL}[q(\theta_s|\wB)||p(\theta_s|\alpha_s)] \\
&+ \mathbb{E}\left[\sum_n \log p(w_n | \theta, \beta) \right],
\end{split}	
\end{equation}
where the expectation is with respect to the variational posterior $q(\theta | \wB)$. 

\dPAM uses stochastic gradient descent to maximize this ELBO to infer the 
variational parameters and learn the model parameters. To finish describing the method,
we must describe how $q$ is parameterized, which we do next. For the subtopic parameters $\beta$,
we learn these using variational EM, that is, we maximize $\mathcal{L}$ with respect to $\beta$.
It would be a simple extension to add a variational distribution over $\beta$ if this was desired.

\paragraph{Re-parameterizing Dirichlet Distribution: } The expectation over the second term in equation \eqref{eq:elbo} is in general intractable and therefore we approximate it using a special type of Monte-Carlo (MC) method \citep{kingma2013auto,rezende2015variational} that employs the re-parametrization-trick \citep{williams1992simple} for sampling from the variational posterior. But this MC-estimate requires $q(\theta|\wB)$ to belong to the location-scale family which excludes Dirichlet distribution. Recently, some progress has been made in the re-parametrization of distributions like Dirichlet \citep{ruiz2016generalized} but in 
this work, following \citet{akash2017} we approximate the posterior with a logistic normal distribution. First, we construct a Laplace approximation of the Dirichlet prior in the softmax basis, which allows us to approximate the posterior distribution using a Gaussian that is in the location-scale family. Then in order to sample $\theta$ from the posterior in the simplex basis we apply the softmax transform to the Gaussian samples. Using this Laplace approximation trick also allows handling different prior assumptions, including other non-location-scale family distributions. 

\paragraph{Amortizing Super-Topics:} As mentioned above, in PAM the super topics need to be sampled for each document in the corpus. This presents a bottleneck in speeding up posterior inference via Gibbs sampling or DMFVI as the number of 
variables to be sampled increases with the amount of data. 
Our use of an amortized inference method allow us
to tackle this bottleneck such that the number of posterior parameters to be learned does not directly depend on the number of documents in the corpus. 
\subparagraph{\dPAM Inference Network}
Recently \citet{akash2017} amortized the cost of learning posterior parameter in LDA with a VAE-type model where they used a feedforward Multi-layer Perceptron (MLP) as the encoder network to generate the parameters for the posterior distribution over the topic proportion vector $\theta$. Like them, we model the posterior $q(\theta_r|\wB)$ as $\mbox{LN}(\theta;f_\mu(\wB),f_u(\wB))$ where $f_\mu$ and $f_u$ are neural networks that generate the parameters for the logistic normal distribution. But their simple encoder cannot be used to learn super-topics because they need to be sampled separately per document, in fact this is one of the reasons \citet{akash2017} assumed the topics to be fixed model parameters.
We now describe our novel structured encoder (inference-network) that can efficiently sample super-topics on-the-fly from a dirichlet prior per document and use them as network weights to generate all the posterior parameters that need to be inferred in PAMs. 

PAM requires a set of variational parameters (topic vectors) per document at each level of the DAG. These topic vectors need to be sampled from a Dirichlet prior $\alpha$ of that level. To generate these parameters, we use one MLP per level which samples the specified number of topic vectors per document for its level.  

Then in order to generate the mixing proportions for the nodes in the lower level, we first note that the Dirichlet distribution is a conjugate prior to the multinomial distribution. This fact can be used to leverage the modern GPU-based computation to generate these mixing proportions since it only involves a dot-product between the mixing proportion from the previous level and the matrix of the topics of the current level. 

Therefore we first stack all the topic vectors of the current level in a 3-D tensor and using a custom implementation for the dot-product 
\footnote{Tensorflow requires that the rank of the tensors in tf.matmul be the same.} 
we generate the mixing proportions for the next lower level. This amortization scheme of our structured encoder gains us significant reduction in training time. We want to point out that the result of above process can also be seen as construction of MLPs on the fly by sampling Dirichlet vectors from our inference networks and stacking them to form weight matrices of the MLPs. This maybe useful in other tasks that require efficient fully Bayesian treatment of the latent variables. 

The decoder in the case of PAM
is similarly just a dot product between the sample from the output distribution of the inference network, the mixing proportions $\theta$ and the sub-topic matrix $\beta$. The only difference is that topics matrix $\beta$ is a global latent variable/model parameter that is sampled only once for the entire corpus.

This framework can be readily extended in several different ways. 
Although in our experiments we always use MLPs to encode the posterior and decode the output, if required other architectures like CNNs and RNNs can be easily used to replace the MLPs. 
As mentioned before, \dPAM can work with non-Dirichlet priors by using the Laplace approximation trick. It can also handle full-covariance Gaussian as well as logistic Normals by simply using the Cholesky decomposition and can therefore be used to learn Correlated Topic Model (CTM) \citep{blei2007correlated}.

At first, the use of an inference network seems strange, as coupling the variational parameters across
documents guarantees that the variational bound will not be as tight.
But the advantage of an inference network is that after the weights of the inference network
have been learned on training documents, we can obtain an approximate posterior distribution
for a new test document simply by evaluating the inference network, without needing
to carry out any variational optimization.
This is the reason for the term \emph{amortized inference}, i.e., the computational
cost of training the inference network is amortized across future test documents.


\subsection{Learning Issues in VAE}
\label{cc}

\begin{figure}
\centering
        \includegraphics[width=0.35\textwidth]{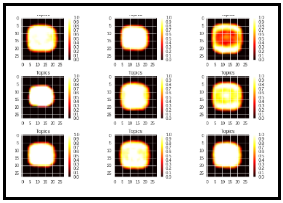}
    \caption{9-randomly sampled "topics" from Omniglot dataset folded back to the original image dimensions. An example of how the topics look like if component collapsing occurs.}
    \label{fig:cc}
\end{figure}

Trained with stochastic variational inference, like VAEs, our PAM models suffer from primarily two learning problems: slow learning and component collapse. 
In this section, we describe each of those problems in more
detail and how we address them.

\subsubsection*{Slow Learning}
Training PAM models even on the recommended learning rate of $0.001$ for the ADAM optimizer \citep{kingma2014adam} generally causes the gradients to diverge early on in training. 
Therefore in practice, fairly low learning rates have been used in VAE-based generative models of text,  which significantly slows down
learning. In this section we first explain one of the reasons for the diverging behavior of the gradients and then propose a solution that stabilizes them, 
which allows training VAEs with high learning rates, making
learning much faster.

Consider a VAE for a model $p(x,z)$
where $z$ is a latent Gaussian variable, $x$ is a categorical variable distributed as $p_{\Theta_d}(x|z) = \mbox{Multinomial}(f_d(z, \Theta_d))$, and the function
$f_d()$ is a decoder MLP with parameters $\Theta_d$ whose outputs 
lie in the unit simplex.
Suppose we define a variational distribution
$q_{\Theta_e}(z|x) = \mathcal{N}(\mu, \exp(u)),$ where $\mu = f_\mu(x, \Theta_\mu)$, $u = f_u(x, \Theta_u)$ are MLPs with parameters $\Theta_e = \{\Theta_\mu,\Theta_u\}$ and $u$ is the logarithm of the diagonal of the covariance matrix.

Now the VAE objective function is
\begin{multline}
\mbox{ELBO}(\pmb \Theta)=-KL[q_{\Theta_e}(z|x) || p(z)] \\+ \mathbb{E}[\log p_{\Theta_d}(x|z)].\label{eq:example-vae}
\end{multline}
Notice that the first term, the KL divergence, interacts only with the encoder parameters.
The gradients of this term $L=KL[q_{\Theta_e}(z|x) || p(z)]$ with respect to $u$ is
\begin{align}
\label{eq:gradu}
\nabla_{u} L &=\frac{1}{2}(\exp(u)-1).
\end{align}

One explanation for the diverging behavior of the gradients lies in the exponential curvature of this gradient. $L$ is sensitive to small changes in $u$, which makes it difficult to optimize it with respect to $\Theta_e$ on high learning rates. 

The
instability of the gradient w.r.t. to $u$ demands an adaptive learning rate for encoder parameters $\Theta_u$ that can adapt to sudden large changes in $\nabla_{u} L$.

We now propose that this adaptive learning rate can be achieved by applying BatchNorm (BN) \citep{ioffe2015batch} transformation to $f_u$. BN transformation for an incoming mini-batch of activations \{$u_{i=1}^m$\} (we overload the notion on purpose here, in general $u$ can come from any layer) is,
\begin{align}
u_{BN}=\gamma \frac{u - \mu_{\text{batch}}}{\sqrt[]{\sigma^2_{\text{batch}+\epsilon}}} + b.
\end{align}
Here, $\mu_{\text{batch}}=\frac{1}{m}\sum_{i=1}^m u_i$, $\sigma^2_{\text{batch}}=\frac{1}{m}\sum_{i=1}^m (u_i - \mu_{\text{batch})^2}$, $\gamma$ is the gain parameter and finally $b$ is the shift parameter. We are specifically interested in the scaling factor $\frac{\gamma}{\sqrt[]{\sigma^2_{\text{batch}}}}$, because the sample variance grows and shrinks with large changes in the norm of the mini-batch therefore allowing the scaling factor to approximately dictates the norm of the activations. Let $L$ be defined as before, the posterior $q$ is now a function of $u_{BN}$. The gradients w.r.t. $u$ and the gain parameter $\gamma$ are
\begin{align}
\label{eq:bn_grad}
\nabla_{u}L &=\frac{\gamma}{\sqrt[]{\sigma^2_{\text{batch}+\epsilon}}}P_u \nabla_{u_{BN}} L \\
\label{eq:bn_gamma}
\nabla_{\gamma}L&=\frac{(u-\mu_{\text{batch}})}{\sqrt[]{\sigma^2_{\text{batch}+\epsilon}}}.\nabla_{u_{BN}} L,
\end{align}
where $P_u$ is a projection matrix. 
If $\nabla_{u_{BN}} L$ is large with respect to the out-going $u_{BN}$, the scaling term brings it down.
\as{I don't understand
the previous sentence. The scaling
term isn't a function of
$\nabla_{u_{BN}} L$, so why would it tend to be small if the gradient
is large? Is there a reason
to think that the two terms are the same order of magnitude?} \cs{they are dependent see (7), basically if $\nabla_{u_{BN}} L$ is large the update to $\gamma$ is small (7) therefore the update to $\nabla_{u} L$ is small (6)}
Therefore, the scaling term works like an adaptive learning rate that grows and shrinks in response to the change in norm of the batch of $u$'s due to large gradient updates to the weights, thus resolving the issue with the diverging gradients. As shown in Figure \ref{fig:kl}, after applying BN to one of the outputs $u$ encoder of the prodLDA model on 20newsgroup dataset \cite{akash2017}, the KL term minimizes fairly slowly (red) compared to the case (blue) when no BN is applied to $u$. We experimentally found that at this point the topics start to improve when the learning rate is $\geq 0.001$. 

In order to establish that the improvement in training comes from the adaptive learning rate property of the gain parameter we replace the divisor in the BN transformation with the $\ell_2$ norm of the activation. We neither center the activations nor apply any shift to them. This normalization performs equivalently and occasionally better than BN, therefore confirming our hypothesis.
\as{This suggests that the exact
type of normalizatoin doesn't matter,
which is good for the reason you say below, but I don't see how
changing the type of normalization confirms the "adaptive learning rate" property.} \cs{not sure about the confusion here.}
It also removes any dependency on batch-level statistics that might be a requirement in models that  make i.i.d assumptions.

\subsubsection*{Component Collapse}

Another well known issue in VAEs such as \dPAM is the problem of
component collapsing \citep{dinh2016training,van2017neural}. 
In the context of topic models, component collapsing is a bad
local minimum of VAEs in which the model only learns a small number of topics out of $K$ \citep{akash2017}. For example, 
suppose we train 
a 3-PAM model on the Omniglot dataset \citep{lake2015human} using the stochastic variational inference from \citet{kingma2013auto}. Figure \ref{fig:cc} shows nine randomly sampled topics for from this model which have been reshaped to Omniglot image dimensions. All the topics look exactly the same, with a few exceptions. This is clearly not a useful set of topics.  

\begin{figure}
\centering
        \includegraphics[width=0.4\textwidth]{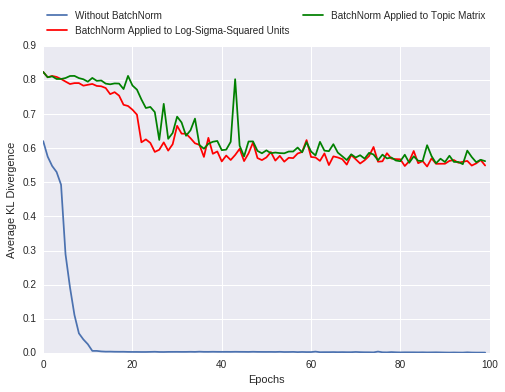}
    \caption{In optimization without any BatchNorm, the average KL gets minimized fairly early in the training. With BatchNorm applied to the encoder unit that produces $\log \sigma^2$, the KL minimization is slow and slower if BatchNorm is also applied to each of the topics in the decoder.}
    \label{fig:kl}
\end{figure}

When trained without applying BN to the $\log \sigma^2$ output of the encoder,
the KL terms across most of the latent dimensions (components of $z$) vanish to zero. We call them collapsed dimensions, since the posterior along them has collapsed to the prior. As a result, the decoder only receives the sampling noise along such collapsed dimensions and in order to minimize the noise in the output, it makes the weights corresponding to these collapsed components very small. In practice this means that these weights do not participate in learning and therefore do not represent any meaningful topic.

Following \citet{akash2017}, we also found that the topic coherence increases drastically when BN is also applied to the topic matrix $beta$ prior to the application of the softmax non-linearity. 
Besides preventing the softmax units to saturate, this slows down the KL minimization further as shown by the green curve in figure \ref{fig:kl}. \as{We've explained the problem
but it's not clear if we've explained the solution.
Does BatchNorm of the topic logits fix component collapsing?}
\as{Did we try l2 normalization instead of batchnorm here?}
\cs{no, by itself it does nothing but if BN,l2 are applied to both places the results are better}

\begin{table*}[h!]
\centering
\caption{Topic coherence
on 20 Newsgroups for 50 topics. PAM models use two super-topics at each level.
For MoLDA, we report separate the coherence for each component in the admixture.}

\label{tab:coherence}

\begin{tabular}{|c|c|c|c|c|c|c|c|c|}
\cline{5-9} \multicolumn{4}{c}{}  &   \multicolumn{5}{|c|}{\textbf{\dPAM}} \\
\hline
\multirow{2}{*}{}        & \multirow{2}{*}{\textbf{\begin{tabular}[c]{@{}c@{}}LDA\\ GIBBS\end{tabular}}} & \multirow{2}{*}{\textbf{\begin{tabular}[c]{@{}c@{}}LDA\\ DMFVI\end{tabular}}} & \multirow{2}{*}{\textbf{\begin{tabular}[c]{@{}c@{}}4-PAM\\ GIBBS\end{tabular}}} & \multirow{2}{*}{\textbf{4-PAM}} & \multirow{2}{*}{\textbf{5-PAM}} & \multirow{2}{*}{\textbf{CTM}} & \multicolumn{2}{c|}{\textbf{MoLDA}} \\ \cline{8-9} 
                         &                                                                               &                                                                               &                                                                                 &                                 &                                 &                               & \textbf{10}      & \textbf{50}      \\ \hline
\textbf{Topic Coherence} & 0.17                                                                          & 0.11                                                                          & 0.20                                                                            & \textbf{0.24}                   & \textbf{0.24}                   & 0.14                          & 0.29             & 0.21             \\ \hline
\end{tabular}
\end{table*}

\begin{table*}[h!]
\centering
\caption{Topic coherence for models trained on 20 Newsgroups dataset for for 100 topics with 50 super-topics.\as{Looks strange for 4-PAM to be highled with MoLDA (50) has better coherence?}}
\label{tab:20news}
\begin{tabular}{|c|c|c|c|c|l|}
\cline{3-6} \multicolumn{2}{c}{}  &   \multicolumn{4}{|c|}{\textbf{\dPAM}} \\
\hline
\multirow{2}{*}{}             & \multirow{2}{*}{\textbf{\begin{tabular}[c]{@{}c@{}}4-PAM\\ GIBBS\end{tabular}}} & \multirow{2}{*}{\textbf{4-PAM}} & \multirow{2}{*}{\textbf{5-PAM}} & \multicolumn{2}{c|}{\textbf{MoLDA}} \\ \cline{5-6} 
                              &                                                                                 &                                 &                                 & \textbf{50}      & \textbf{100}     \\ \hline
\textbf{Topic Coherence}      & 0.19                                                                            & \textbf{0.22}                   & 0.21                            & 0.24             & 0.21             \\ \hline
\textbf{Training Time (Min.)} & 594                                                                             & 11                              & 16                              & \multicolumn{2}{c|}{16}             \\ \hline
\end{tabular}
\end{table*}

\begin{table*}[h!]
\centering
\caption{Topic coherence for models trained on NIPS dataset for 50 topics with 2 super-topics.}
\label{tab:robust_nips}
\begin{tabular}{|c|c|c|c|c|l|}
\cline{3-6} \multicolumn{2}{c}{}  &   \multicolumn{4}{|c|}{\textbf{\dPAM}} \\
\hline
\multirow{2}{*}{}        & \multirow{2}{*}{\textbf{\begin{tabular}[c]{@{}c@{}}4-PAM\\ GIBBS\end{tabular}}} & \multirow{2}{*}{\textbf{4-PAM}} & \multirow{2}{*}{\textbf{5-PAM}} & \multicolumn{2}{c|}{\textbf{MoLDA}} \\ \cline{5-6} 
                         &                                                                                 &                                 &                                 & \textbf{10}      & \textbf{50}      \\ \hline
\textbf{Topic Coherence} & 0.033                                                                           & \textbf{0.042}                  & 0.039                           & 0.036            & 0.024            \\ \hline
\end{tabular}
\end{table*}
 
\begin{table*}[h!]
\centering
\caption{Topic coherence for models trained on NIPS dataset for 100 topics with 50 super-topics.}
\label{tab:robust}
\begin{tabular}{|c|c|c|c|c|l|}
\cline{3-6} \multicolumn{2}{c}{}  &   \multicolumn{4}{|c|}{\textbf{\dPAM}} \\
\hline
\multirow{2}{*}{}             & \multirow{2}{*}{\textbf{\begin{tabular}[c]{@{}c@{}}4-PAM\\ GIBBS\end{tabular}}} & \multirow{2}{*}{\textbf{4-PAM}} & \multirow{2}{*}{\textbf{5-PAM}} & \multicolumn{2}{c|}{\textbf{MoLDA}} \\ \cline{5-6} 
                              &                                                                                 &                                 &                                 & \textbf{50}      & \textbf{100}     \\ \hline
\textbf{Topic Coherence}      & \textbf{0.047}                                                                                 & 0.041                           & 0.045                  & 0.025            & 0.024            \\ \hline
\textbf{Training Time (Min.)} & 892                                                                                & 19                              & 26                              & \multicolumn{2}{c|}{11}             \\ \hline
\end{tabular}
\end{table*}

\section{Experiments and Results}
\label{exp}

We evaluate how \dPAM inference performs for 
different architectures of PAM models when compared to the state-of-art collapsed Gibbs inference. To this end we evaluate three different PAM architectures, 4-PAM, 5-PAM and MoLDA, on two different datasets, 20 Newsgroups and NIPS abstracts \citep{Lichman:2013}. 
We use these two data sets because they 
represent two extreme settings. 20 Newsgroups is a large dataset (12,000 documents) but with a more restricted 
vocabulary (2000 words) whereas the NIPS dataset is smaller in size (1500 abstracts) 
dataset but has a considerably larger vocabulary (12419  words). 
We compare inference methods both on time required for training
as well as topic quality.
As a measure of topic quality, 
we use the topic coherence metric (normalized point-wise mutual information), which as shown in \citet{lau2014machine} corresponds very well with human judgment on the quality of topics. We do not report perplexity of the models  because 
it has been repeatedly shown to not be a good measure of topic coherence and even to be negatively correlated with the topic quality in some cases \citep{lau2014machine,chang2009reading,akash2017}. 

We start by comparing the topic coherence across the different topic models on the 20 Newsgroup dataset. 
 We train an LDA model using both collapsed Gibbs sampling\footnote{We used the Mallet implementation \citep{mallet}.} \citep{griffiths2004finding} and Decoupled Mean-Field Variational Inference (DMFVI)\footnote{We used the \texttt{scikit-learn} implementation \citep{scikit-learn}.} \citep{blei2003latent}. Using Mallet, we train a 4-PAM model using 10000 iterations of collapsed Gibbs sampling and using \dPAM we train a 4-PAM, a 5-PAM, a MoLDA and a correlated topic model (CTM). In this experiment we use 50 sub-topics for all models.
{For MoLDA we use two
 mixture components with 10 and 50 topics.}
 For 4-PAM and 5-PAM, we use two super-topics following \citet{li2006pachinko}, and two additional
 super-duper-topics for 5-PAM.
 \as{We haven't defined super-duper-topic --- also this word sounds a bit
 silly to my ear, maybe better two say two topics at the top level, two super topics at the next level.}
Results are shown in Table \ref{tab:coherence}.
All PAM models perform better than LDA-type models,
showing that more complex PAM architectures do improve the quality of the topics. 
Additionally 4-PAM and 5-PAM models trained on \dPAM beat all the LDA models for topic quality. MoLDA and CTM trained using \dPAM also perform competitively with the LDA models but the CTM model falls significantly behind PAM models on topic coherence. 

Next, to study the effect of increasing the number of  PAM supertopics,
we increase the number of super-duper-topics to 10, super-topics to 50 and sub-topics to 100. \as{It's unfortunate that we don't have at least one of the LDA models for this; it would be nice to see whether the
quality of PAM increases faster than LDA as the number of parameters increases. Also remember the reviewer suspicion argument if you leave an obvious comparison out without good reason.} Table \ref{tab:20news} shows the topic coherence for each of these models and also the training time. Not only our inference method produces better topics it also is an order of magnitude faster than the state-of-art Gibbs sampling based inference for 4-PAM. Note that we run the sampler for a total of 3000 iterations with the burn-in parameter set to 2000 iterations. \as{The reviewers will want to know how these numbers are chosen, because they are key to the running time comparison. Could we have gotten better performance with fewer iterations?}

For the NIPS dataset, we repeat the same experiments only for the PAM models again under the same exact settings as described above. Reported in Table \ref{tab:robust_nips} are the topic coherence for smaller PAM models with 50 sub-topics. Again, we allowed 10,000 Gibbs iterations which took more than a day to finish but did not beat \dPAM-trained models on topic quality. For the bigger PAM models we replicated the experiments from the original paper. As reported in Table \ref{tab:robust}, we found that while the collapsed Gibbs based 4-PAM model produced the best topics,
it did so in 15 hours. On the other hand, \dPAM-trained models produced topics with comparable quality with a fraction of the inference time,
because this method is
able to leverage the GPU architecture for computing dot-products very efficiently. \as{There might be an amortization benefit as well?} We are not aware of GPU-based implementations of other inference method for PAMs.
\as{Why don't we report training time for Tables 1 and 3? Wouldn't it be more consistent if we did?}

\subsection{Hyper-Parameter Tuning}

For the experiments in this section we did not conduct extensive hyper-parameter tuning. We used 
a grid search for setting the encoder capacity according to the dataset. As a general guideline for PAM models, the encoder capacity should grow with the vocabulary size. For the learning rate, we used the default setting of $1e^{-3}$ for the Adam optimizer for all the models. We used a batch size of 200 for 20 Newsgroups  as used in \citep{akash2017} and 50 for the NIPS dataset. 
We found that the topic coherence, especially for the smaller NIPS dataset, is sensitive to the batch size setting and initialization. For certain settings we were able to achieve higher topic coherence than the average topic coherence reported in this section.

\section{Related Work}
\label{related}
Topic models have been explored extensively via directed \citep{blei2003latent,li2006pachinko,blei2007correlated,blei2004htm} as well as undirected models or restricted Boltzmann machines \citep{larochelle2012neural,hinton2009replicated}. Hierarchical extensions to these models have received special attention since they allow capturing the correlations between the topics and provide meaningful interpretation to the latent structures in the data. 

Recent advancements in blackbox-type inference method \citep{kucukelbir2016automatic,ranganath2014black,mnih2014neural,khan2017conjugate} have made it easier to try newer models without the need of deriving model-specific inference algorithms.

\section{Conclusion}
\label{con}
In this work we introduced \dPAM, which extends the idea of variational inference in topic models via structured VAEs. We found that the combination of amortized inference and modern GPU software allows for an order of magnitude improvement in training time compared to standard inference mechanisms in such models. We hope that this will
allow future work to explore new and more complex architectures for deep topic models.
 



\bibliography{naacl}
\bibliographystyle{acl_natbib}

\end{document}